# A NOVEL TOPOLOGY DESIGN APPROACH USING AN INTEGRATED DEEP LEARNING NETWORK ARCHITECTURE

**Sharad Rawat and M.-H. Herman Shen** [1]
Department of Mechanical and Aerospace Engineering
The Ohio State University
201 W 19th Avenue
Columbus, OH 43210

## ABSTRACT

Topology design optimization offers tremendous opportunity in design and manufacturing freedoms by designing and producing a part from the ground-up without a meaningful initial design as required by conventional shape design optimization approaches. Ideally, with adequate problem statements, to formulate and solve the topology design problem using a standard topology optimization process, such as SIMP (Simplified Isotropic Material with Penalization) is possible. In reality, an estimated over thousands of design iterations is often required for just a few design variables, the conventional optimization approach is in general impractical or computationally unachievable for real world applications significantly diluting the development of the topology optimization technology. There is, therefore, a need for a different approach that will be able to optimize the initial design topology effectively and rapidly. Therefore, this work presents a new topology design procedure to generate optimal structures using an integrated Generative Adversarial Networks (GANs) and convolutional neural network architecture. The discriminator in the GANs as well as the convolutional network are initially trained through the dataset of 3024 true optimized planner structure images generated from a conventional topology design approach (SIMP). The convolutional network maps the optimized structure to the volume fraction, penalty and radius of the smoothening filter. Once the GAN is trained, the generator produced a large number of new unseen structures satisfying the design requirements. The corresponding input variables of these new structures can be evaluated using the trained convolutional network. The structures generated by the GANs are also minutely post-processed to aberrations. Validation of the results is made by generating the images with same conditions using existing topology optimization algorithms.  This paper presents a proof of concept which will lead to further scale research efforts of amalgamating deep learning with topology design optimization.

---

[1] Corresponding Author, email: shen.1@osu.edu

## 1. INTRODUCTION

Topology optimization [1-3], a branch of design optimization, is a mathematical method to solve a material layout problem constrained for a given design domain, loading, and boundary conditions. This method determines the optimal distribution of material such that the structure has desired properties (e.g. minimizing the compliance of structure) while satisfying the design constraints. The majority of topology optimization applications, in automotive and aerospace industries for structural, acoustics, fluid, and thermal related fields [4], presently use Simplified Isotropic Material with Penalization (SIMP). This algorithm penalizes the intermediate density values (non-binary) during iterations for better convergence. Section 2 discusses about SIMP model in detail.

Topology optimization is transforming the design processes. However, it also brings a few drawbacks to the design process. One of the major drawbacks is the heavy computational costs incurred for a large design problem. There have been many studies to improve the topology optimization algorithms and reduce the computational costs of this method. Kim and Yoon [5] discusse a multi-resolution design strategy to design progressively from low to high resolution during the optimization process. To reduce the computational power required by the conventional topology algorithms, a new convergence criterion was proposed by Kim et al. [6] where the design variables were reduced during optimization. Liu et al [7], proposed an unsupervised machine learning method to reduce the dimensionality of design variables using the K-means clustering algorithm.

With the advent of advanced graphics processing units (GPU) for general purpose computing and recent developments in artificial neural networks, deep learning is emerging in large range of industries including speech recognition, medical diagnosis, autonomous vehicles, and e-commerce. Artificial Neural Networks, specifically deep learning networks, provide a capability to map a highly non-linear relationship between the variables by learning through the data. Sosnovik and Oseledets [8] applied a deep convolutional network for speeding up the design process. This converted the topology optimization process from a primarily finite element problem to an image segmentation task. However, this study lacked the design information and settings necessary for topology optimization. A very recent breakthrough in the form of generative models has led to a further application of deep learning in design processes. A major class of generative networks is GANs [9]. GANs have been widely regarded as the most promising invention in the last decade in the field of deep learning. GANs have many variations like a super-resolution GAN which enhances the resolution of the images improving the details in an image, changing the environment of the images (Cycle GAN) and many more.

Recently Yu et al [10], [11] proposed using generative modeling techniques for generating new designs. A Variational Autoencoder (VAE) was used for generation of new structures. They extended their work by using GAN for improving the design, hence enhancing the lower resolution to a higher resolution. However, GAN was not used for the generation of new structures. Li et al [12] proposed a new architecture to generate a new structure from a combination of GAN and multiple convolutional networks.

In this study, we propose a model using an integrated GANs and convolutional neural network architecture to generate new structures using limited data samples. A convolutional neural network is then trained to map the structures with their respective design conditions. Section 2 discusses the conventional method used for the problem statement of topology optimization. The concepts and architectures of the neural networks along with the dataset used for training purpose used in this article are discussed in section 3. Section 4 elaborates the results of the study and the corresponding discussions.  Section 5 concludes this article with the inferences, potential future work and potential issues while using GANs. Validation of the results is made by generating the images with same conditions using SIMP topology optimization algorithm.  This work displays a successful application of GANs for design process by generating sub-optimal planar structures without using the existing algorithms repeatedly.

## 2. PROBLEM STATEMENT

Topology optimization is a modern optimization algorithm which aims to distribute material inside a given design domain [13]. Concepts of finite element method along with optimizing techniques like genetic algorithm, method of moving asymptotes, optimality criteria and level set method are utilized in this algorithm. Mathematically, it is an integer optimization problem where each finite element of the discretized design domain constitute a design variable. This design variable represents the density of that element which takes a value of either 0 (with no material) or 1 (with material), hence named a binary density variable at times. This problem can be relaxed by using converting the discrete integer problem into a continuous variable problem namely, SIMP. In SIMP, non-binary solutions are penalized with a factor *p*. Mathematically, the topology optimization implementation suing SIMP can be defined as:

$$\min_{x} : \qquad c(x) = U^T K U = \sum_{e=1}^{N} (x_e)^p \; u_e k_o u_e$$

$$subjected\ to: \qquad \frac{V(x)}{V_0} = f,$$

$$\mathrm{KU} = \mathrm{F},$$

$$0 < x_{min} \leq x \leq 1$$

Where c(x) is the objective function, U and F are the global displacement and force vectors respectively. K is the global stiffness matrix, $u_e$ and $k_o$ are the element displacement vector and stiffness matrix, respectively, x is the vector of design variables, $x_{\min}$ is a vector of minimum relative densities (non-zero to avoid singularity), N (= nelx×nely) $N = \nabla_x \times \nabla_y$ is the number of elements used to discretize the design domain, p is the penalization power (typically p = 3), V (x) and V0 is the material volume and design domain volume, respectively and f (volfrac) is the prescribed volume fraction [14][15]. This existing method is used to generate a dataset of 3024 samples as discussed in Section 3.

## 3. A NOVEL INTEGRATED DEEP LEARNING NETWORK

A Wasserstein Generative Adversarial Network (WGAN) is used as a generative model where there is a minimax game between a generator and a discriminator. The generator is trying to fool the discriminator and the discriminator attempts to identify fake images. In the proposed model, the discriminator is first trained through the dataset of true images followed by generated images from the generator. It is different from conventional GANs in their objective function. WGAN uses earth mover distance (EM) instead of Jenson-Shannon Divergence in GANs. In this paper WGAN is coupled with a convolutional neural network. The convolutional network is also trained using the dataset of 6048 samples. These 6048 samples are generated from existing dataset of 3024 samples using data augmentation techniques. The convolutional network maps the optimized structure to the volume fraction, penalty and radius of the smoothening filter. Once the WGAN is trained, the generator produced new unseen structures. The corresponding input variables of these new structures can be evaluated using the trained convolutional network. The conceptual flow of the proposed method is shown in Fig. 1. The networks' architectures are discussed in detail later in the section.

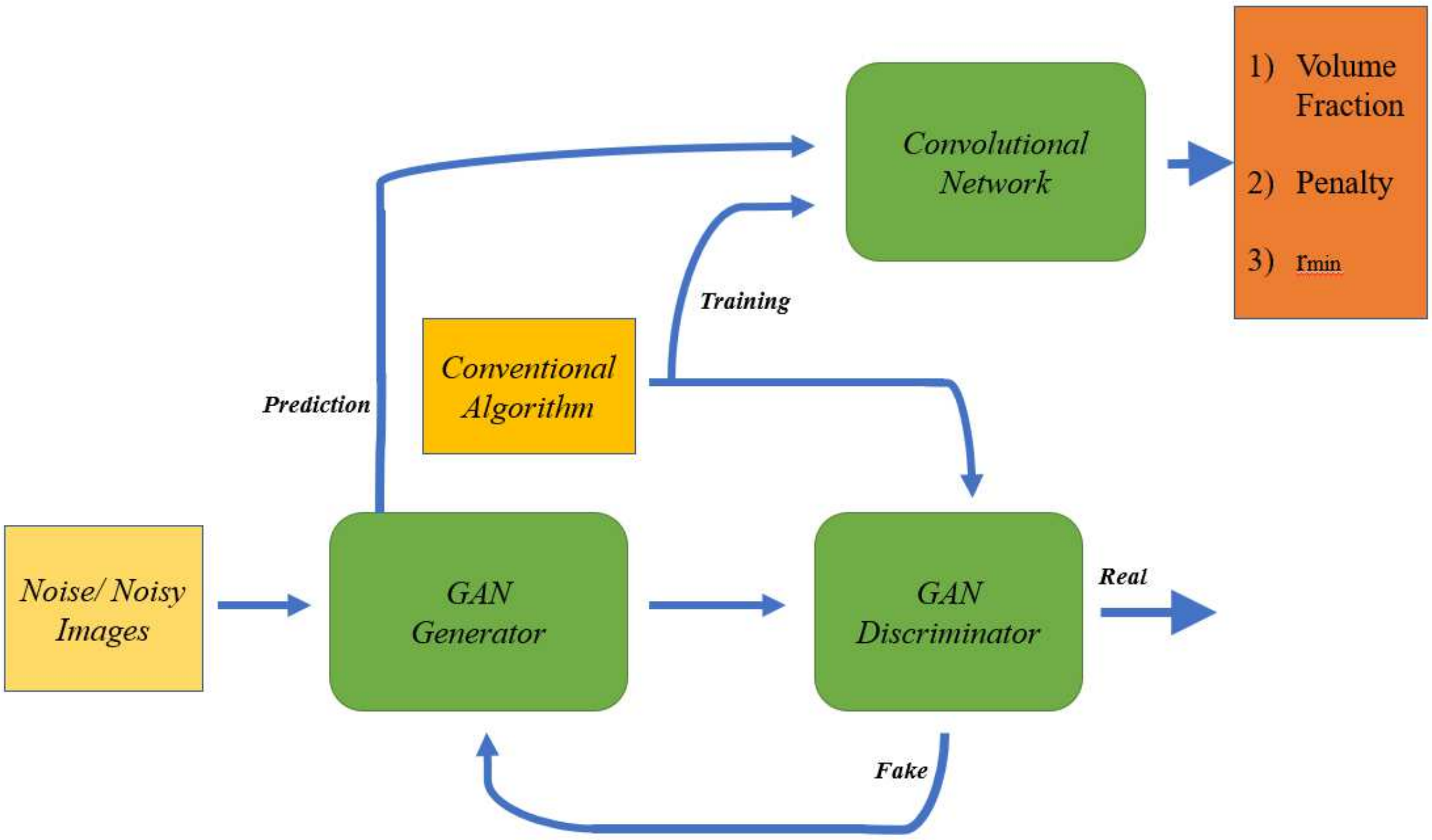

Fig. 1 Conceptual flow of the proposed method

**Convolutional Neural Network**

Convolutional neural networks (CNN) are deep artificial neural networks that are primarily used for image classification, cluster them into classes or perform a regression mapping data. Lecun [15], proposed an architecture called LeNet. It is one of the first CNN architectures. CNNs have been proved to be efficient while working with images. Fig. 2 depicts the architecture of CNN used in this study. This CNN maps the topology optimized structures in the form of images to their corresponding design conditions i.e. volume fraction (vol_frac), penalty (penal) and radium of smoothening filter ($r_{min}$). In this architecture, an input of 120x120x1 is fed into the structure followed by layers of convolution, max pooling and batch normalization. After 3 such layers of each, a branch is extracted which outputs a volume fraction. Volume fraction is extracted out early in the architecture because it's features, depicting volume fraction (presence of material), are very apparent and can be learnt with less complex architectures. The main branch is fed to another set of layers composed of convolution, max pooling and batch normalization. Penal is extracted after this layer and $r_{min}$ is obtained after evaluated after passing through more hidden layers. Since this is a regression problem, no activation function is used in output layers and 'ReLU' is used for all the remaining layers. The loss function for the regression is mean squared error. Adam optimizer with learning rate of 0.0002 is used for the study. Adam is preferred over other optimizers because of the continuous decay of learning rate. The total number pf parameters are 1,206,306. The architecture was created using Tensorflow backend Keras library in python 3.5.

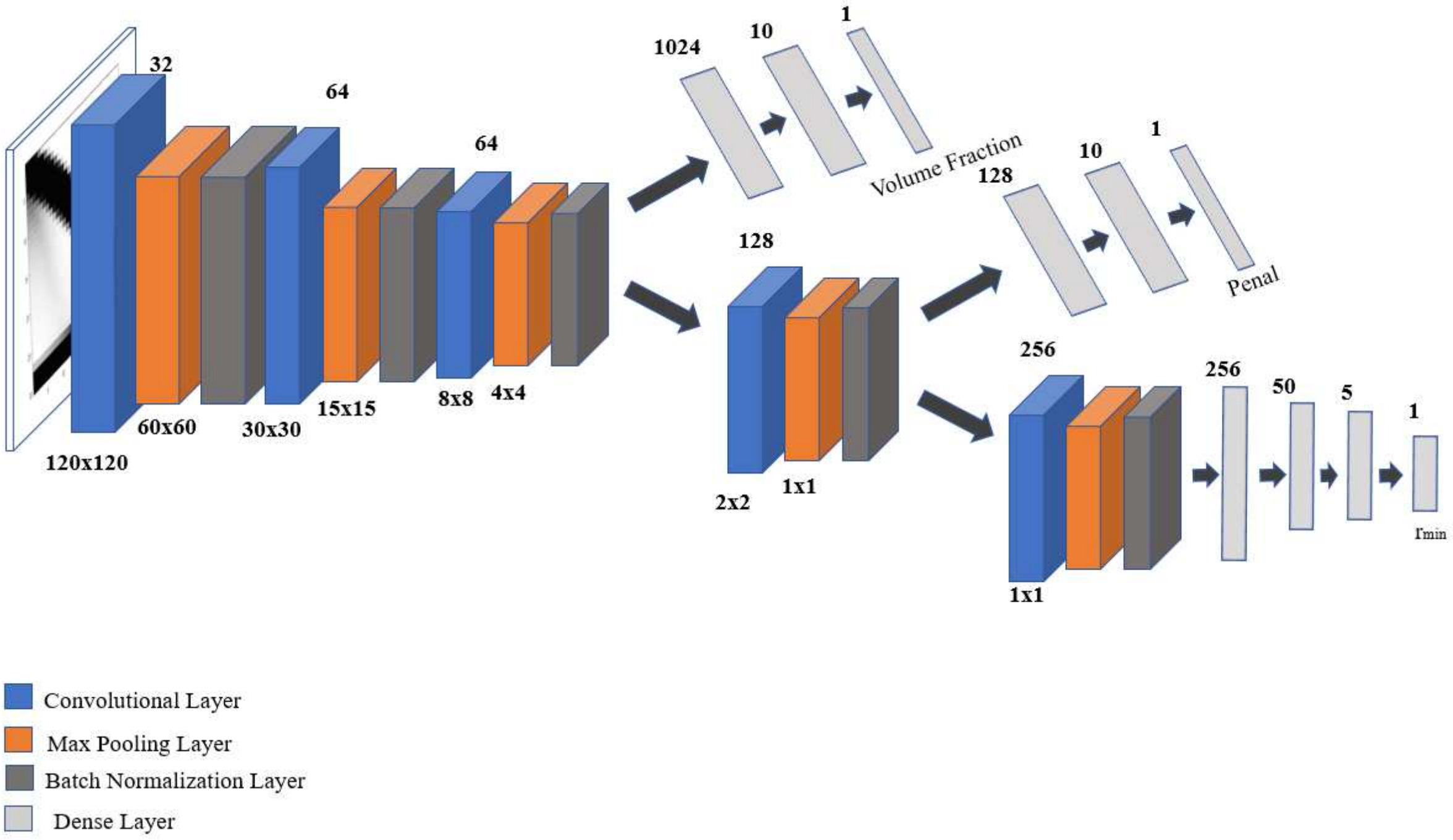

Fig. 2 Architecture of the convolutional neural networks

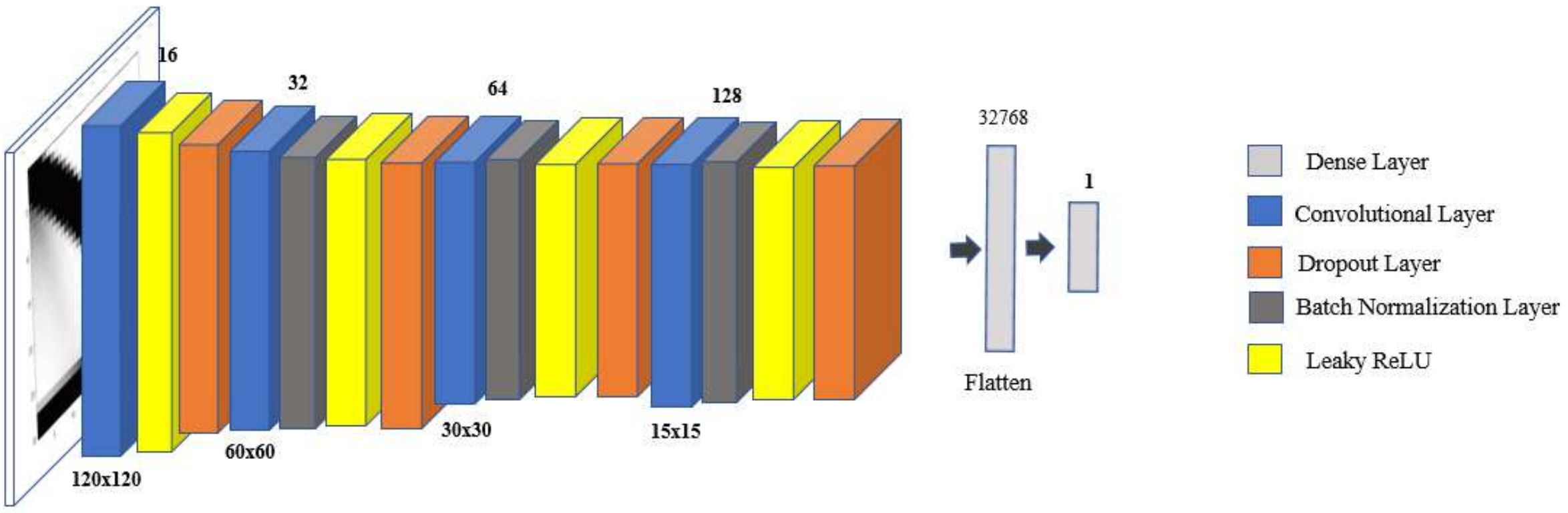

Fig.3 Architecture of discriminator of the GAN used

### Generative adversarial network

Generative adversarial Network (GAN) has shown a promising performance for image generation. GANs consists of 2 networks, a generator G(z) and a discriminator D(x). These two networks are adversaries of each other during training process. The discriminator is trained to successfully distinguish between the true and fake images. Whereas, the generator is trained to generate realistic image from noise such that the discriminator is unable to distinguish between the real and fake images. Mathematically, the objective function is described as:

$$\min_{G}\max_{D} V(D,G) = E_{x\sim P_{data}}[\log(D(x)] \;+\; E_{z\sim P_z}[\log(1-D(G(z)))]$$

Where x is the image from training dataset $P_{data}$ and z is the noise vector sampled from $P_z$.

There have been certain modifications to the vanilla GAN algorithm to produce good quality images. Firstly, a deep convolutional GAN (DCGAN) has proved to be producing realistic images. In an attempt to stabilize the training, Wasserstein GAN (WGAN) is used where the objective function is modified and the weights are clipped to enforce the Lipchitz constraint.

The architecture for WGANs used for this study is shown in Fig. 3 and Fig. 4. This architecture consists of convolutional neural networks in generators as well as discriminator. The generator is deeper than the discriminator. No pooling layer was either in discriminator or the generator [17]. For the discriminator, the input size is a tensor of 120x120x1. After convolutions and batch normalization after every layer, the output of the discriminator is a 1x1 tensor between -1 (real images) and 1 (fake images). For the generator, the input is a noise of latent dimension of size 100x1. After repeated convolutions layers and batch normalization layers, the output from the generator is a tensor of size 120x120x1. The discriminator and the generator are alternatively trained. RMSProp is used as the optimizer with learning rate of 0.00005. An activation function of Leaky ReLU with alpha of 0.2 is used. To avoid unstable training, an attempt has been made to clip the weights of discriminator and to smoothen the labels on both sides i.e. the labels were changed to (-0.9, 0.9) instead of (-1, 0) [18].

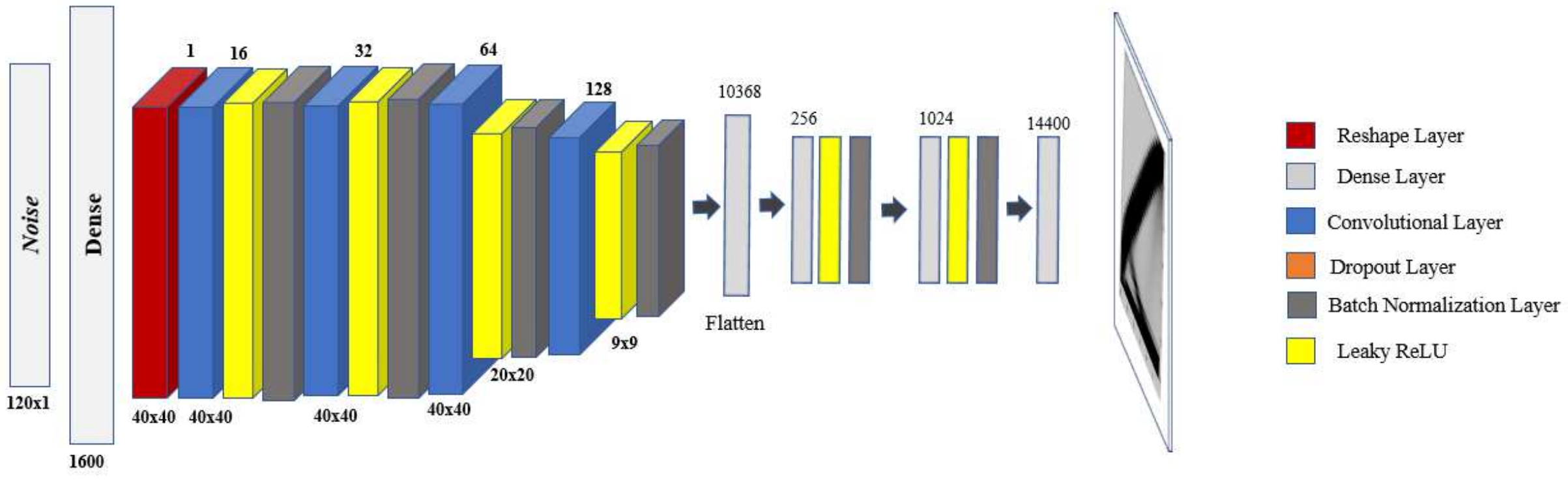

Fig. 4 Architecture of the generator used in GAN

**Dataset**

The training data is computed using the conventional topology optimization code which uses the SIMP model. A dataset of 3024 samples was generated using volume fraction (vol_frac), penalty (penal) and radius of the smoothening filter ($r_{min}$) as design variables. Following is bounds of each variable:

| | | |
|---|---|---|
| 1) vol_frac – 0.3-0.8 | 2) Penal – 2-4 | 3) $r_{min}$ – 1.5-3 |

There can be more variables to this problem but since this article is a proof of concept, only 3 scalar variables are chosen. Each dataset was created at a resolution of 120 X 120 pixels with same boundary condition of a cantilever beam loaded at the mid-point of hanging end. The dataset generated has 2 purposes:

1) Train the GAN.
2) 2) Train the convolutional network.

Since for training of the convolutional network, this dataset was insufficient, the dataset of 3024 samples was, hence, augmented to increase the number of samples to 6048. Augmentation of the data samples was done by adding noise at random locations for each sample. Hence the data has 2*3024 samples for training of convolutional networks.

## 4. RESULTS and DISCUSSIONS

In this study, generative adversarial networks are used to construct optimal design structures for a given boundary condition and optimization settings. A conventional topological optimization algorithm is used to generate a small number of datasets as displayed by Figure 5. Samples taken from the datasets are supplied to the discriminator network in GAN for training. An adversarial game between the discriminator and the generator is performed which aids the generator to learn the distribution of dataset $P_{data}(x)$.

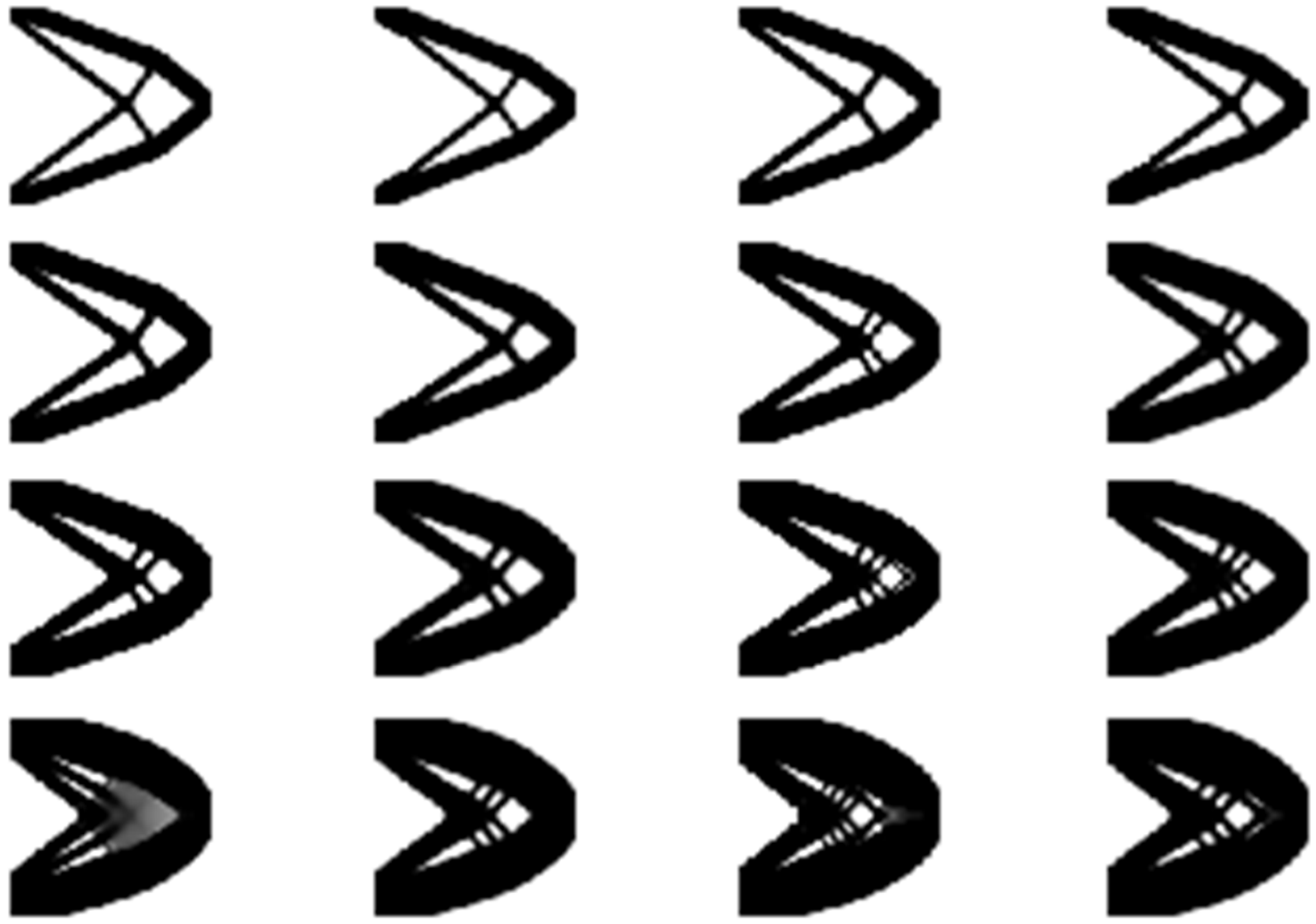

Fig.5 Optimal structures from conventional algorithm

A convolutional neural network (CNN) is trained on the same dataset from conventional algorithm. This CNN maps the optimized structure to its corresponding optimization settings. After the completion of training of GAN, samples from GAN are fed to this CNN to obtain the optimization settings. Figure 6 shows the samples generated from GAN after training. It can be seen that although, some of the structures generated by GAN are not defined clearly, GANs have been able to replicate the distribution of dataset ($P_{data}$). Since there are a mix of volume fractions, it can be said that this GAN is not subjected to mode collapse.

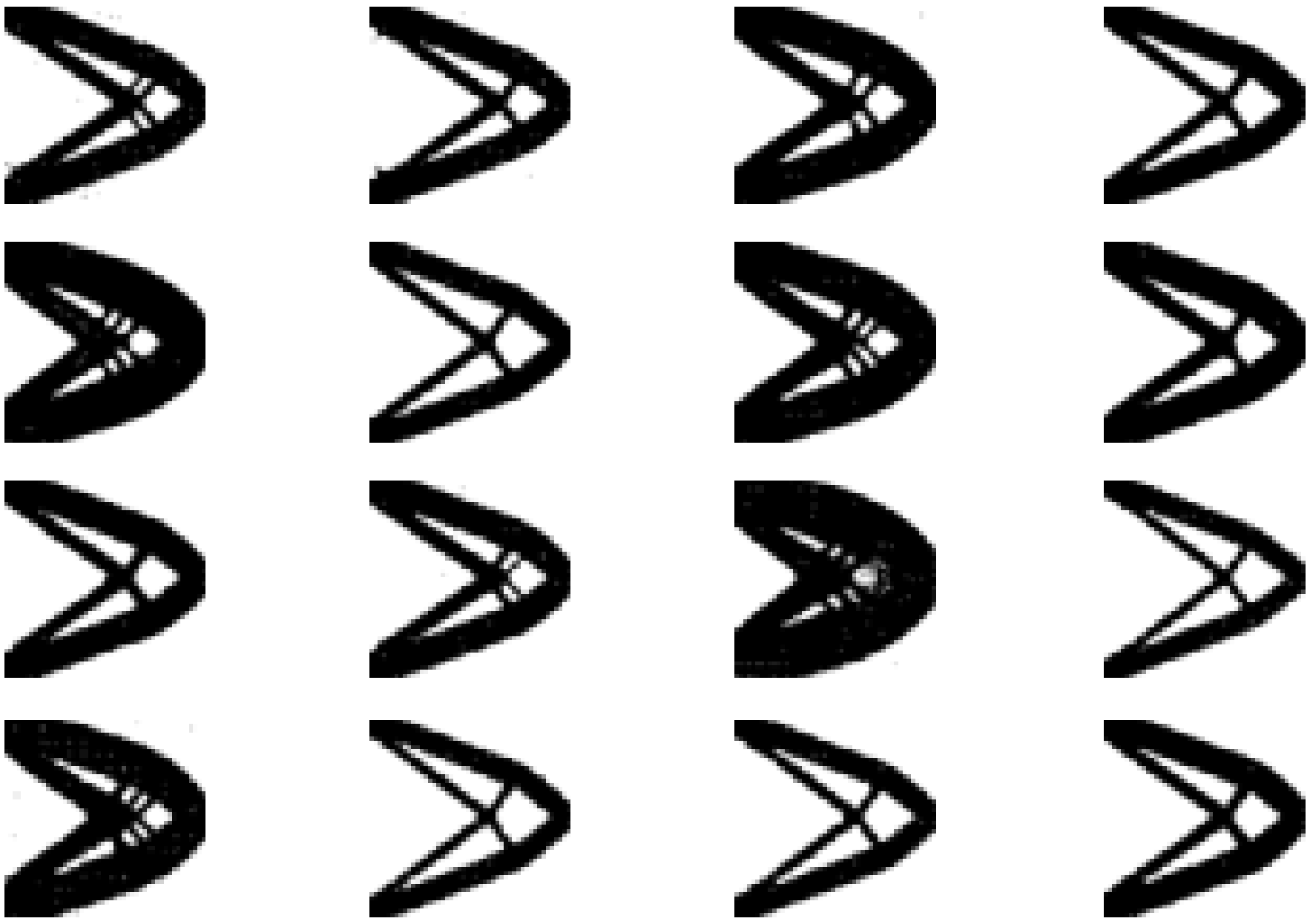

Fig. 6 optimal structures generated from GAN

To validate the performance of GAN, a qualitative evaluation is conducted comparing the structures from GAN against the structures from conventional algorithm with the same optimizations setting. Table 1. presents the comparative results. It can be observed that the structures generated from GAN are extremely close to their corresponding structures from the conventional algorithm.

**Post-processing**

Planar structures from GANs are post-processed in order further improve the quality of the images. A threshold filter is created where all the pixels above a value of 0.5 are rounded to 1 and pixel values below 0.5 are rounded to 0. This filter improves the quality of the structure but the resulting structure is extremely sharp. To further smoothen the planar structure, a Gaussian filter with a kernel of 5 was applied on the structure. The resulting structure is in good agreement as can be observed in the Table 1.

Table 1. Comparative results from GAN and the corresponding conventional algorithm

| Generated Structure from WGAN | Design parameters from Convolutional Network | Topology optimized structure conventional algorithm |
|---|---|---|
| 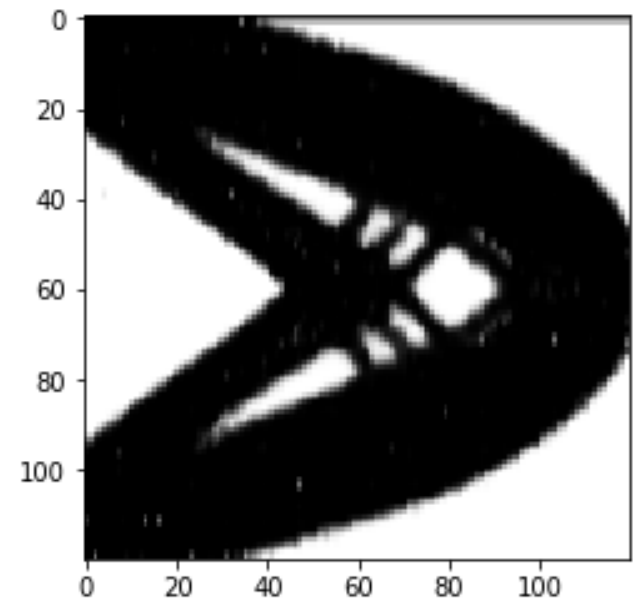 | vol_frac = 0.607<br>Penal = 3.0857<br>$r_{min}$ = 1.769 | 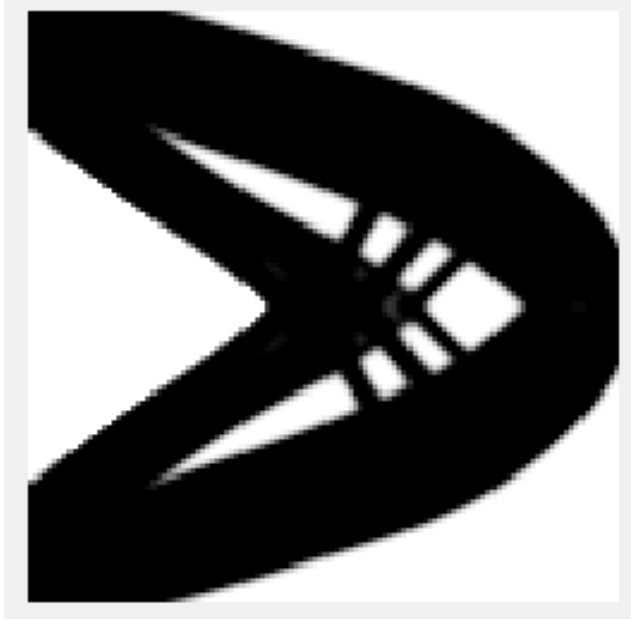 |
| 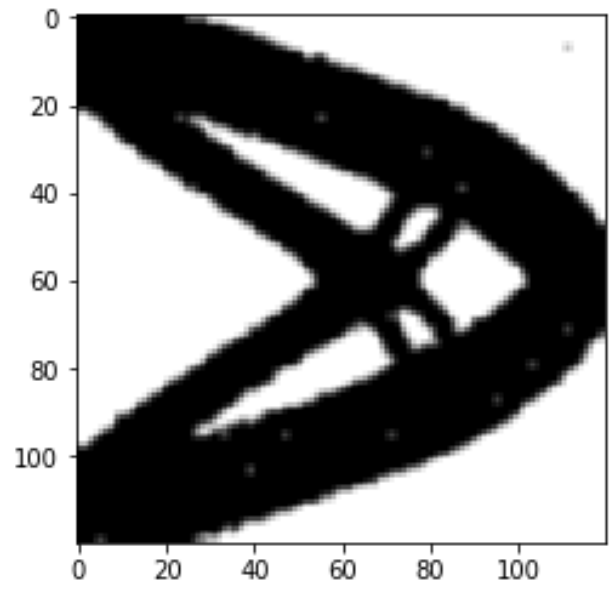 | vol_frac = 0.551<br>Penal = 3.607<br>$r_{min}$ = 2.592 | 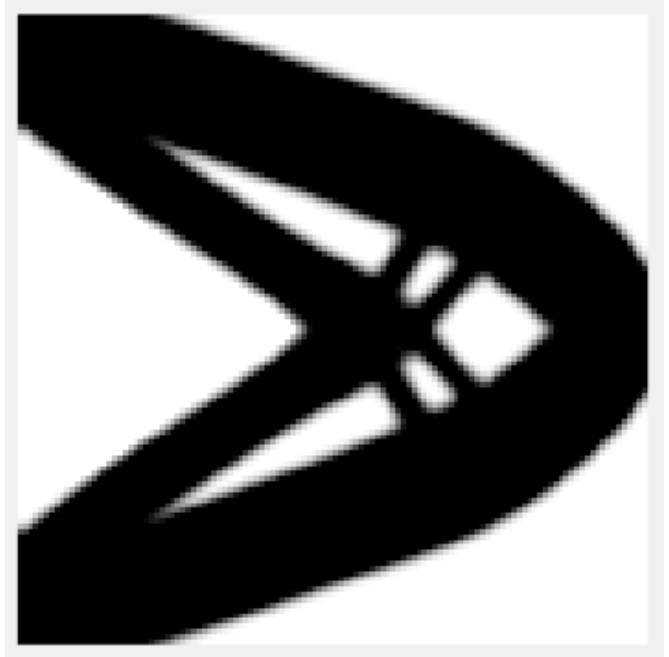 |
| 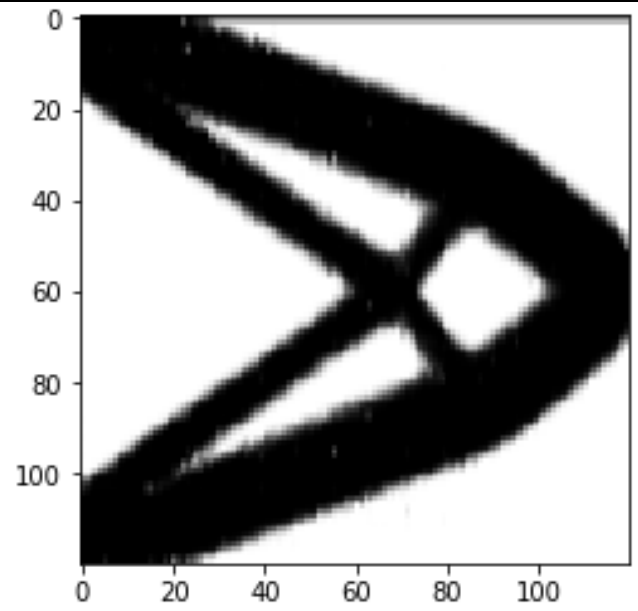 | vol_frac = 0.451<br>Penal = 2.861<br>$r_{min}$ = 1.699 | 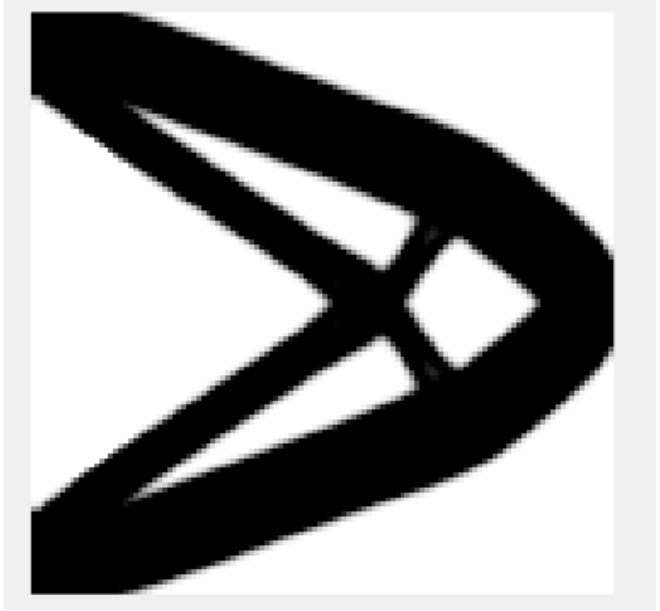 |
| 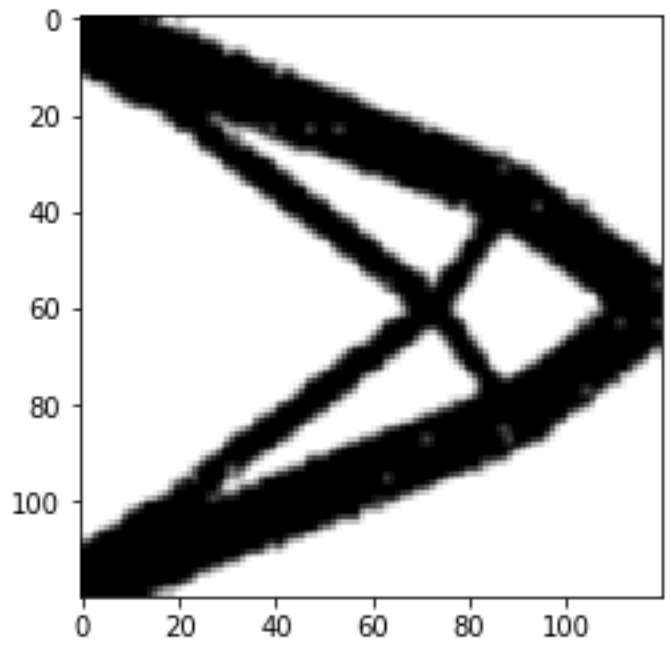 | vol_frac = 0.295<br>Penal = 3.301<br>$r_{min}$ = 2.736 | 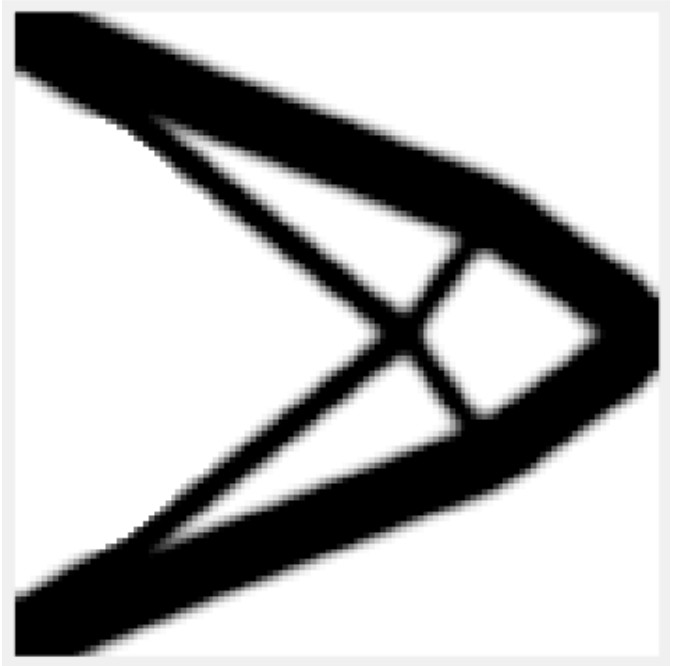 |

## 5. CONCLUSION

In this study, GANs are to generate optimal structures for a certain set of constraints and conditions. The structure produced through GAN is mapped to its correspond optimization conditions utilizing the convolutional neural network. On comparing the quality of structures against the structures generated from existing algorithm, GANs show a capability of generating a sub-optimal topology once trained with a very small dataset of 3024 samples. With the good agreement of the GAN results against the conventional algorithm, GANs can reconstruct models for further complex model and optimization settings which are computationally expensive. By using a deep learning model in design, designers will be able to quickly perform design iterations for a variety of conditions. GANs also provide an infinite supply of data of optimal structures which can be used for other deep learning models. Nevertheless, several critical improvements need to be addressed to implement the technology in topology optimization successfully. These include the extension to conditional GANs and the relaxation of the boundary condition constraint instead of the fixed cantilever used in this study. Despite the aforementioned future works, this work presents the initial attempts made to amalgamate deep learning with mechanical design processes. Culmination of this work is expected to engender a new comprehensive optimal design technology for topology optimization. The technology can be expected to provide expand and enable a flexible and scalable topology design process in real world applications.